\documentclass[runningheads]{llncs}
\usepackage[style=numeric, sorting=none, maxbibnames=2]{biblatex}
\usepackage{amsmath,amssymb,amsfonts}
\usepackage{booktabs}
\usepackage{subcaption}
\usepackage{float}
\usepackage{orcidlink}
\usepackage{array}
\usepackage{colortbl, xcolor}
\usepackage{makecell}
\usepackage[a4paper,margin=3.5cm]{geometry}

\usepackage[T1]{fontenc}
\usepackage{graphicx, verbatim}
\newcommand{\absdiv}[1]{%
  \par\addvspace{.5\baselineskip}
  \noindent\textbf{#1}\quad\ignorespaces
}

\begin{document}
\title{Native space based pipelines outperform template space based pipeline in subcortical segmentation}
\titlerunning{Subcortical segmentation: Native vs Template space}
%
\author{Tomás Lima\inst{1}\orcidlink{https://orcid.org/0009-0002-1895-9028} \and
Daniel Novák\inst{1}\orcidlink{https://orcid.org/0000-0001-6996-9344} \and
Eduard Bakštein\inst{1,2}\orcidlink{https://orcid.org/2222--3333-4444-5555}}
\authorrunning{T. Lima et al}
%
\institute{\textsuperscript{1} Department of Cybernetics, Czech Technical University in Prague, Prague, Czech Republic\\ 
\textsuperscript{2} National Institute of Mental Health, Klecany, Czech Republic}

\maketitle              
\begin{abstract}
\absdiv{Introduction} Accurate segmentation of subcortical regions is critical for neurosurgical planning and functional research. Most automated methods rely on template space coregistration, which may compromise patient-specific accuracy, particularly in small structures. We identify a need to evaluate whether native space approaches offer a measurable advantage, which we evaluate in the context of movement disorders.

\absdiv{Methods} We developed two UNet-based segmentation pipelines of the Subthalamic Nucleus (STN) -- a common surgical target in Parkinson's Disease -- and the neighbouring Red Nucleus (RN) and Substantia Nigra (SN). We collected 7T and 3T MRI data from five public datasets. The pipelines were evaluated in the native-space against manual labels. We further investigated the effect of the template resolution. Motivated by the hypothesis that models may better learn target boundaries in higher field, we tested the transferability of 7T-trained models to 3T clinical images, and whether synthetic 3T training data -- generated via a disentangled representation learning method -- could help bridging this domain gap.

\absdiv{Results} On held-out 7T data, the native pipeline consistently outperformed the template one. For the STN, native-space Dice reached 0.775 ± 0.055 versus 0.713 ± 0.051 (1 mm template), with HD95 of 0.79 ± 0.24 mm versus 1.17 ± 1.10 mm, respectively. Similar advantages were observed for the RN and SN. Increasing template resolution did not improve accuracy. When applied to 3T images, all models showed a considerable performance drop. Adding synthetic 3T data yielded only modest improvements, though without degrading 7T performance.

\absdiv{Conclusion} Native-space segmentation is preferable for applications requiring patient specific anatomical fidelity, such as the surgical planning in PD. Bridging the 7T-to-3T domain gap remains an open challenge, motivating future work on domain adaptation tailored to subcortical structures.
\end{abstract}


\section{Introduction}
Segmentation of subcortical nuclei is a recurrent clinical practice that has applications in many clinical scenarios, as well as in functional organization research. This region of the brain is well-captured through Magnetic Resonance Imaging (MRI), which has become the standard pre-surgical planning step in different medical procedures, generating also valuable research data \cite{mri_best, serranova2019topography}. The level of tissue detail achieved by MRI is, among other factors, highly dependent on the strength of the scanner magnetic field. Briefly, stronger fields (7T and above) achieve better contrast between the tissues visualized, which in turn allows for the capture of higher resolution images \cite{heim2017magnetic, kim2019automatic}. In the particular case of subcortical structures, their clear identification is particularly important in motor-related neuropathologies, like Parkinson's Disease (PD), as they play an important role in motor control pathways \cite{b6, b7}, and are a common target for neurosurgical interventions.

Despite the advantages of higher-field MRI, low-field scans (1.5-3T) are still the common clinical practice, rendering the limited contrast subcortical structures difficult to identify. For this reason, many automatic subcortical segmentation tools perform indirect targeting, involving the warping of a pre-acquired anatomical or histological atlas to the patient's image. This brings advantages in terms of human and time resources needed, but lacks specificity over anatomical disparities \cite{daisne2013atlas}, which are very prone to occur in subcortical nuclei \cite{variability_4}. Existing evidence, including \textcite{variability_1} and \textcite{variability_2} suggests this variability is too high to warrant satisfactory results with atlas based segmentation.

The majority of existing tools for subcortical segmentation perform co-registration of the patients' MRI images to a standardized template space \cite{manjon_pbrain_2020, varga2025precise}, such as the Montreal Neurological Institute Space (MNI) \cite{DBS_non_linear_deformations}. This step allows for consistent anatomical reference across subjects, making it easier to locate smaller, low-contrast structures based on population-averaged coordinates \cite{template_advantages}. However, this normalization has one major drawback: the surgical intervention is done in patient's native space, making it necessary to transform the segmentations obtained in the template space back to the native space.

In this work, we investigate the potential negative effect that the back and forth registration to template space may have on segmentation performance in a neurological application. Even though this issue was given some attention in the past \cite{crivello2002comparison, krishnan2006accuracy}, the research interest has been limited until more recently, when new evidence appeared due to its clinical implications \cite{pijar2025modeling, giorgi2022locus}. However, quantified evaluation on subcortical segmentation is still missing.

Based on 7T public data, we tackled the segmentation of PD associated subcortical nuclei, the Subthalamic Nucleus (STN), and the close neighboring structures Red Nucleus (RN) and Substantia Nigra (SN). To this end, we created two segmentation methods both based on deep learning U-Net model: one relying on the MNI template space, and the other operating directly in the native space. We further analyzed the influence of the MNI template resolution on the results. We additionally investigated if the knowledge about the target structures' boundaries and shapes learned in the clearer 7T data could be directly transferable to lower field 3T clinical images. Finally, we hypothesized if the addition of synthetic 3T data generated from the original 7T images would improve this knowledge transferability. We trained and evaluated the pipelines on a dataset curated from publicly available 7T data. 

In summary, we aim to answer the following four research questions:
\begin{itemize}
    \item[Q1] Does subcortical segmentation directly in native space provide more accurate results than a segmentation pipeline using co-registration to template space?
    \item[Q2] What is the effect of template space resolution on the answer of Q1?
    \item[Q3] Is knowledge directly transferable from a 7T-trained subcortical segmentation model to 3T clinical scenario?
    \item[Q4] Does data augmentation using synthetic images improve the knowledge transferability assessed in Q3?
\end{itemize}


\section{Materials and Methods}
The following chapter describes the Template and Native segmentation pipelines implemented. Both segmentation pipelines (hereafter methods) work upon the same preprocessing of the MRI data, and both rely on the nnUNet framework \cite{nnunet}. Next, we introduce the two techniques used to generate the synthetic 3T images. We start by describing the public data collected for this work and the evaluation metrics used.


\subsection{MRI Data}
We collected T2w images -- which are the standard sequence providing higher contrast in basal ganglia segmentation \cite{abosch2010assessment} -- from five open source datasets. Since T1w images were available, these were also collected. 7T images were collected from the Open Science CBS Neuroimaging Repository (CBS)\cite{CBS}, the \emph{Atlasing of the basal ganglia} (ATAG)\cite{atag}, and a recently published dataset (Chu 2025) \cite{paired_beijing}. 3T and 7T images were collected from a 3T and 7T paired dataset (Paired) published by \textcite{paired}. Independent 3T images of 10 subjects were collected from the OASIS repository \cite{oasis}. Despite also having paired 3T and 7T images, the 3T data of Chu 2025 was not included in this study, as the contrast of the structures of interest was too low to allow manual labeling with sufficient accuracy. A summary of the collected data can be found in Table \ref{tab1}. Due to the restrictions of currently available 7T datasets, the age range of the collected data does not match the typical age of PD patients. However, we believe our segmentation pipeline findings to be generalizable to different age and condition groups.

\begin{table}
\centering
\caption{Specifications of the five datasets used in this study. Resolution is isotropic, unless specified otherwise.}
\setlength{\tabcolsep}{3pt}
\begin{tabular}{c c c c c c}
\toprule 
Dataset& 
Subjects&
Age&
Field-strength&
Resolution (mm)&
Train/Test\\
\midrule
CBS& 20&  $26.8\pm4.2$ &7T&0.5&15/5\\
\rowcolor[gray]{.9}
ATAG& 25&  $26.9\pm9.4$ &7T&0.5&15/5\\

Chu 2025& 20&  $23.4\pm1.4$ &7T&0.4×0.4×1.0&15/5\\
\rowcolor[gray]{.9}
Paired& 9 & $32.1\pm5.6$  &3T, 7T&\begin{tabular}[c]{@{}c@{}}
0.8 (3T)\\
0.65 (7T)
\end{tabular} &
\begin{tabular}[c]{@{}c@{}}
0/9 (3T)\\
7/2 (7T)
\end{tabular} \\

OASIS& 10&  N/A  &3T&1&0/10 \\

\bottomrule
\end{tabular}
\label{tab1}
\end{table}


\subsubsection{Train and Test Split}
We used a split validation procedure. The balance of approximately 80\% training to 20\% test was kept for each dataset, with the exception of the OASIS dataset and the Paired 3T subset, which were both used fully in the test set to obtain validation on unseen data source. Table \ref{tab1} includes the contribution of each dataset for both the training and the test subsets. 

\subsubsection{Labeling procedure}
All T2w images were manually segmented by a trained expert using the ITK-Snap software \cite{itk-snap}. All labels were generated in the native space.

\subsubsection{Evaluation metrics}
Segmentation accuracy was assessed using two complementary metrics. The Dice similarity coefficient (Dice) quantifies the volumetric overlap between two segmentations, ranging from 0 (no overlap) to 1 (perfect agreement). The 95th-percentile Hausdorff distance (HD95) measures the largest boundary discrepancy between two surfaces after discarding the top 5\% most extreme distances, thus providing a robust assessment of surface-level accuracy that is less sensitive to outliers \cite{huttenlocher2002hausdorff}.

\subsection{Preprocessing of MRI images}
The T2w echoes of the subjects in CBS and ATAG datasets were averaged for each voxel to generate a single T2w image. Due to the high variation introduced by the different sources used in this study, z-normalization was applied to the image intensities separately for the 3T and 7T subsets. All images were skull-stripped using the SynthStrip tool \cite{synthstrip}.

For the ATAG dataset only, each patient's T2w image was affine-registered to the T1w image. In this dataset, this step was used for all MNI related operations, since the ATAG's T2w images consisted only of slab captures, not full-brain images, and their direct co-registration to the template space did not achieve sufficient accuracy.

\subsection{Data Augmentation and UNet Training} \label{nnunet_training}
All the UNet models trained in this work followed the same data augmentation and training setup. Data augmentation was set to rotation (max 10º), translation (max 10 mm), and scaling (max 10\%). 
The augmented training datasets were used to train 3D full-resolution convolutional networks within the nnUNet v2 framework \cite{nnunet} in a 5-fold cross validation scenario, using the latest UNet Residual Encoder presets \cite{nnunet_new_presets}. Training employed a custom nnU-Net trainer derived from the default nnU-Net training pipeline. Models were trained for 15 epochs, and mirror-based data augmentation was disabled. All models were trained and tested in a workstation equipped with Python 3.9.2, an NVIDIA GeForce RTX 3090 GPU (24 GB VRAM) and CUDA 12.3.

\subsection{Space Normalization Based Model - Template Method}
The automatic segmentation template method relies on the space normalization of all the images to the MNI Nonlinear Asymmetric version of 2009 brain template \cite{mni}. In order to study the effect of the template characteristics on the final segmentation output, two MNI versions were selected: one with a lower resolution of 1~mm (MNI 152 NLIN 2009c) - Template Method (1mm MNI) - and another at high-resolution of 0.5mm (MNI 152 NLIN 2009b) - Template Method (0.5~mm MNI). Both MNI templates were skull-stripped. The transformation matrix of each subject' T2w sequence to the T2w version of the template was computed, with exception of the ATAG dataset -- here, the T1w sequences were used to compute the registration to the T1w version of the MNI spaces, and the computed transformation matrices were then applied to the T2w images and label maps of all subjects. All the transformations in this step were affine. The model could thus be trained in the MNI space, and the stored transformation matrices allowed for the inverse registration of the model output back into the native spaces for evaluation.

\begin{figure}[!t]
\centerline{\includegraphics[width=1\columnwidth]{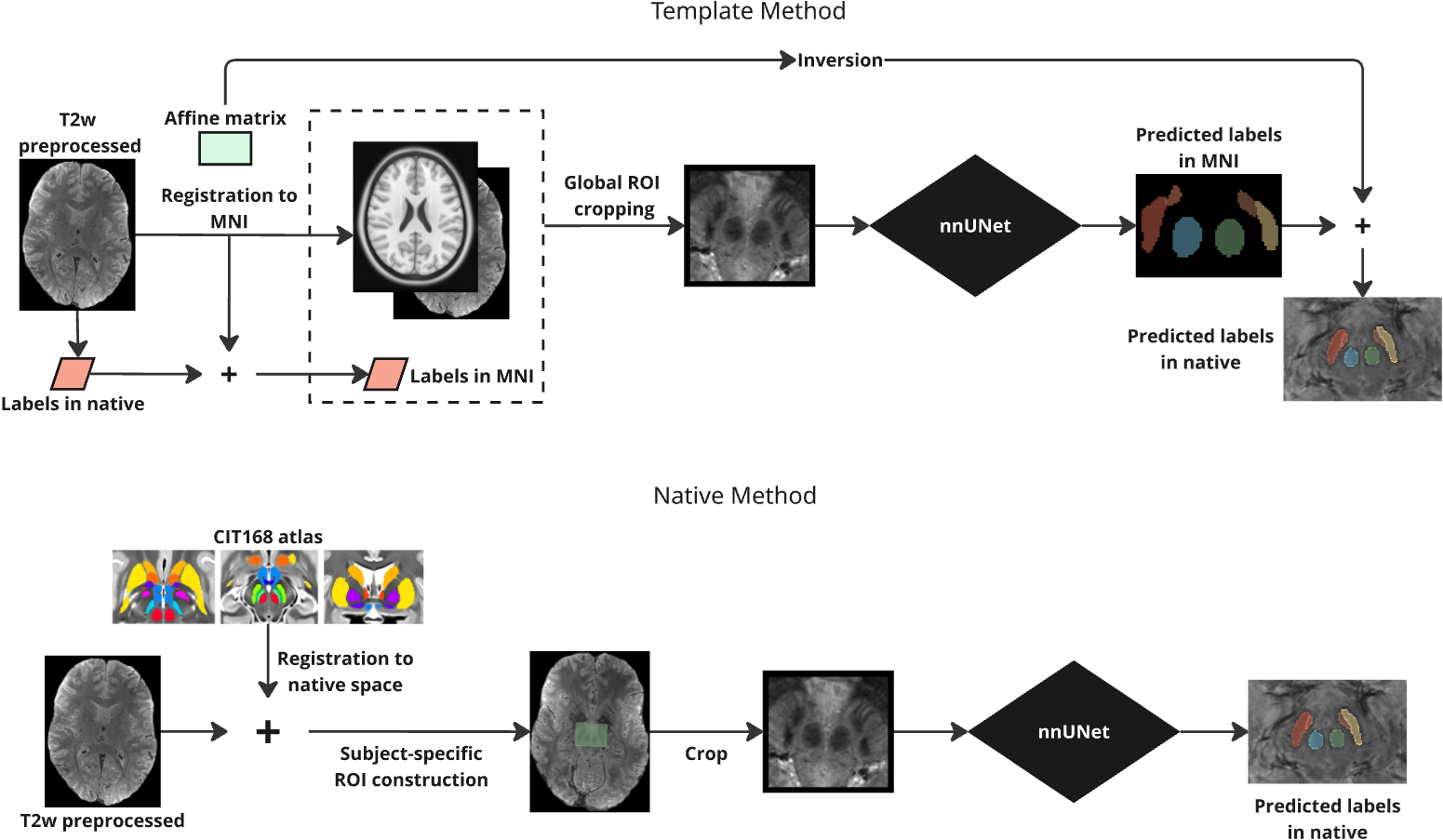}}
\caption{Diagrams of the automatic segmentation pipelines implemented: Template-space based method (top), which requires the alignment (affine transform) of brain images to MNI standard space and the inverse transformation of the output label maps back to native space; Native-space based method (bottom), operating fully in the native space and relying on the registration of the CIT168 atlas to the native space only for the construction of the subject-specific ROI around the target structures.}
\label{fig:pipelines}
\end{figure}

The overall pipeline of this method is illustrated in the top diagram of Figure \ref{fig:pipelines}.
With all data aligned in the MNI space, a common Region Of Interest (ROI) was defined by storing the maximum and minimum coordinates of labeled voxels in the MNI space across all subjects, and adding a margin of 3 and 5 voxels, for the 1~mm and 0.5~mm variations, respectively.

The resulting dataset was augmented and used to train a UNet as detailed in Section \ref{nnunet_training}. The output segmentations were inverse-transformed to the native space, following the method described by \textcite{inverse_transform}, where non grid-point positions are derived by linear interpolation.

\subsection{Native Space Based Model - Native Method}
First, subject-specific ROIs were created using the probability maps of the CIT168 atlas \cite{cit168}, which focuses on subcortical structures in an age group similar to the collected dataset, with successful previous applications to basal ganglia segmentation\cite{cit168_success, MNI_distortion_and_CIT168_success}. We used the version of the atlas with isotropic resolution of 1~mm. 
For ROI generation, the probabilistic labels of the target structures of this study (RN, SN and STN) were thresholded at 0.3 and converted to binary maps. A cubic ROI mask surrounding the atlas selected labels was created, and a 3 voxel margin was added in every direction. Since the atlas is built on top of MNI, the template's affine registration to every subject T2w image was computed, and then applied to the constructed cubic ROI. Each subject's images and labels were cropped following the resulting subject-specific ROI. 

Note that, in opposition to what happens in the Template method, in this case it is the MNI that is being warped to the subjects' native space, with the single goal of constructing the subject-specific ROI. 

In order to obtain the same dimensions of the images to be input to the model, all were cropped to a shape of (90, 80, 60) based on an analysis of the coordinates of the label maps after augmentation, guaranteeing that the target structures were always included with a safety margin. Finally, the data augmentation and UNet training described in Section \ref{nnunet_training} was followed. The overview of the Native segmentation pipeline is represented in the bottom diagram of Figure \ref{fig:pipelines}.

\subsection{Generation of 3T synthetic data}
One of the questions that this work aimed at answering, was to understand if the knowledge about the target structures'  learned in the more well-defined 7T images could be transferred to lower-field clinic-like 3T images (Q3). We hypothesized that synthetic 3T images, generated from the training 7T images, could enhance this possible transferability (Q4). Here, we describe two methods tested for the generation of the synthetic images.

\subsubsection{Simple - Histogram Standardization}
To serve as a baseline, a simple method was created. Here, the synthetic images were obtained by downsampling the high-resolution 7T images to 1mm voxel size and applying a histogram matching technique. Piecewise-linear histogram matching happens in two steps: (i) the 3T training images are used to compute intensity percentiles and building a standardized reference histogram; (ii) the 7T images' intensities are mapped to the standard histogram. This mapping follows the alignment suggested by \textcite{nyul1999standardizing} and was applied using the TorchIO package \cite{perez-garcia_torchio_2021}. The 3T standard histogram was generated by averaging the normalized intensity histograms of the 3T images of the Paired and OASIS datasets.

\subsubsection{Advanced - MURD}
\textcite{MURD} introduced a multi-site unsupervised representation disentangler (MURD) framework for the harmonization of mutli-site 3D medical images. MURD consists of a content-style disentangled cycle translator, composed of site-specific style-encoders and style-generators, and a content-encoder, generator, and discriminator shared by all-sites. In summary, MURD disentangles the content (anatomical) and style (site-dependent intensity variations) features and is then able to generate a new image based in the extracted content features and some target style features. We have adapted MURD in order to apply it to a 3T-7T image translation task.

Since MURD can work with unpaired images, we treated 3T and 7T images of different datasets as being two different sites. For the training of MURD, we used the 3T images of the Paired and OASIS datasets, and the Paired, CBS and Chu 2025 datasets were used as sources of the 7T images. In total, 19 3T and 25 7T images were used for training MURD. All the available 3T images available in the study were used, and a balanced collection of the 7T image datasets was selected, in order to make MURD adaptable to the different image sources.

After the initial preprocessing, the images were resliced to a 0.8mm isometric resolution, and cropped into (256, 256, 192) shape. Z-normalization was also applied. We kept the 2.5D nature of the original MURD framework (the model has three input channels, which cover three adjacent 2D slices), and trained the model for 15 epochs, until sufficient convergence was achieved. Once the model was trained, it was fed with all the 7T images of the training subset, and their 3T synthetic versions were stored. The target style was generated using the trained 3T-specific style generator. After inference, the resulting synthetic images were downsampled to 1~mm isotropic resolution.


\section{Results}
All models were evaluated on held-out test data containing both 7T and 3T acquisitions. Performance is reported for the three studied structures (RN, SN and STN) using Dice and HD95. Quantitative results are summarized in Table \ref{tab:q1_q2} (7T test) and Table \ref{tab:q4} (3T test), while Figures \ref{fig:q1_q2_plot}-\ref{fig:segmentation_examples} provide distributional and qualitative comparisons. Additionally, shape-related errors are presented in the Appendix figures \ref{fig:vol_area_method_I_1mm} and \ref{fig:vol_area_method_II_native}. 

\subsection{Q1 - Template vs. Native segmentation}

\begin{figure}[!t]
\centerline{\includegraphics[width=1\columnwidth]{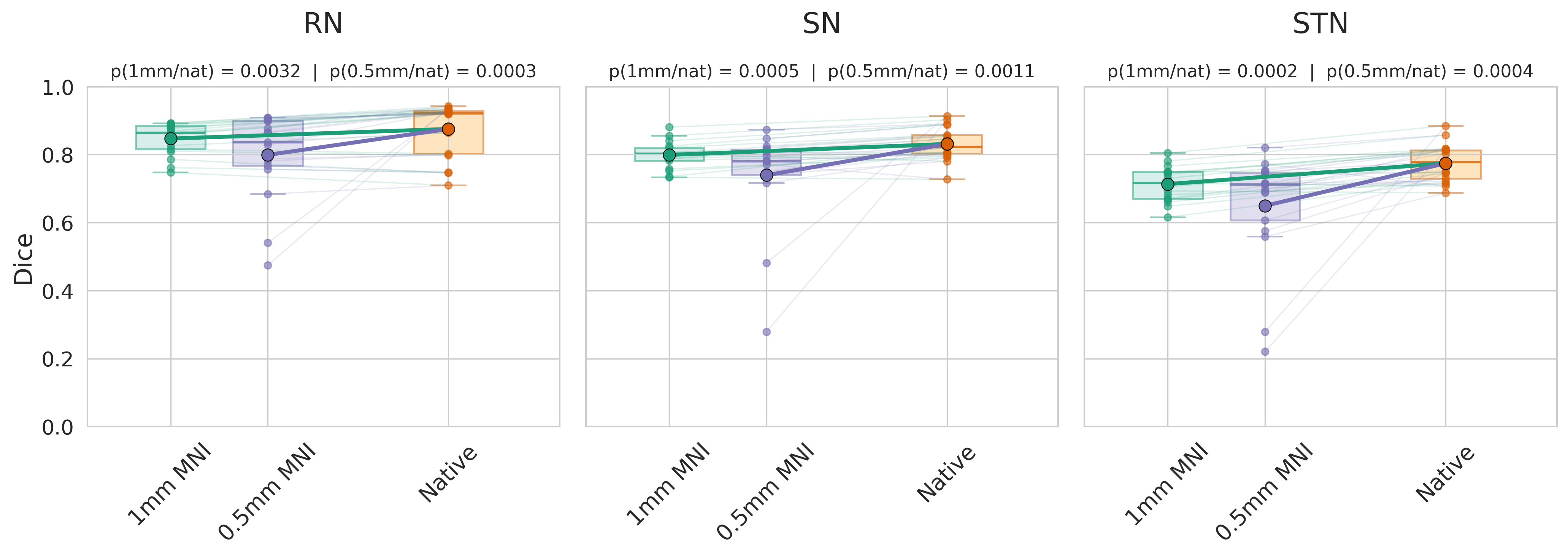}}
\caption{Dice scores comparison between both Template method variants (using 1~mm and 0.5~mm template resolution) and Native method on the 7T test subset, trained with the 7T train set data only.}
\label{fig:q1_q2_plot}
\end{figure}

To answer Q1, we evaluated both methods, trained only with 7T data, in the 7T test subset. In this scenario, the Native-space pipeline achieved the highest Dice scores (Table \ref{tab:q1_q2}, Figure \ref{fig:q1_q2_plot}), outperforming both Template-space variants. The improvement was most pronounced for the STN, our main target and also the smallest and lowest contrast structure in the set, where the Native method reached Dice 0.775 ± 0.055, compared to 0.713 ± 0.051 (Template 1~mm) and 0.650 ± 0.166 (Template 0.5~mm). A similar pattern holds for RN and SN, where Native pipeline also obtained the best Dice (RN: 0.875 ± 0.080; SN: 0.831 ± 0.049).

HD95 showed the same general trend of better boundary accuracy by the Native method, yielding markedly improved HD95 for SN and STN relative to template-space processing (Table \ref{tab:q1_q2}). The RN was the only structure in which the Template method (1~mm) achieved better HD95, but the difference is minimal -- 0.63 ± 0.16 against 0.66 ± 0.23 of the Native method.

Figure \ref{fig:segmentation_examples} qualitatively supports these findings: on 7T images, all methods produce plausible contours, but native-space outputs adhere more tightly to the manual labels, especially in case of the STN.

\begin{table}
\begin{center}
\caption{Segmentation performance on the 7T test set for models trained with 7T data only. Best results per structure are in bold.}
\label{tab:q1_q2}
\begin{tabular}{llcc}
\toprule
Structure & Pipeline & Dice & HD95 (mm) \\
\midrule
RN & Template (1mm) & 0.847 ± 0.047 & \textbf{0.63 ± 0.16} \\
\rowcolor[gray]{.9}
RN & Template (0.5mm) & 0.799 ± 0.128 & 0.86 ± 0.59 \\
RN & Native & \textbf{0.875 ± 0.080} & 0.66 ± 0.23 \\
\hline
SN & Template (1mm) & 0.799 ± 0.040 & 0.94 ± 0.90 \\
\rowcolor[gray]{.9}
SN & Template (0.5mm) & 0.739 ± 0.146 & 1.08 ± 0.64 \\
SN & Native & \textbf{0.831 ± 0.049} & \textbf{0.72 ± 0.22} \\
\hline
STN & Template (1mm) & 0.713 ± 0.051 & 1.17 ± 1.10 \\
\rowcolor[gray]{.9}
STN & Template (0.5mm) & 0.650 ± 0.166 & 1.15 ± 0.59 \\
STN & Native & \textbf{0.775 ± 0.055} & \textbf{0.79 ± 0.24} \\
\bottomrule
\end{tabular}
\end{center}
\end{table}

\subsection{Q2 - Influence of resolution in Template segmentation}
Increased template space resolution negatively affected the performance of the Template method. The 0.5~mm pipeline showed lower Dice and higher spread across the tested structures than the 1~mm MNI version (e.g., STN Dice drops from 0.713 ± 0.051 at 1~mm to 0.650 ± 0.166 at 0.5~mm -- Table \ref{tab:q1_q2}).

\subsection{Q3 - Knowledge transferability to 3T images}
\begin{figure}[!t]
\centerline{\includegraphics[width=0.5\columnwidth]{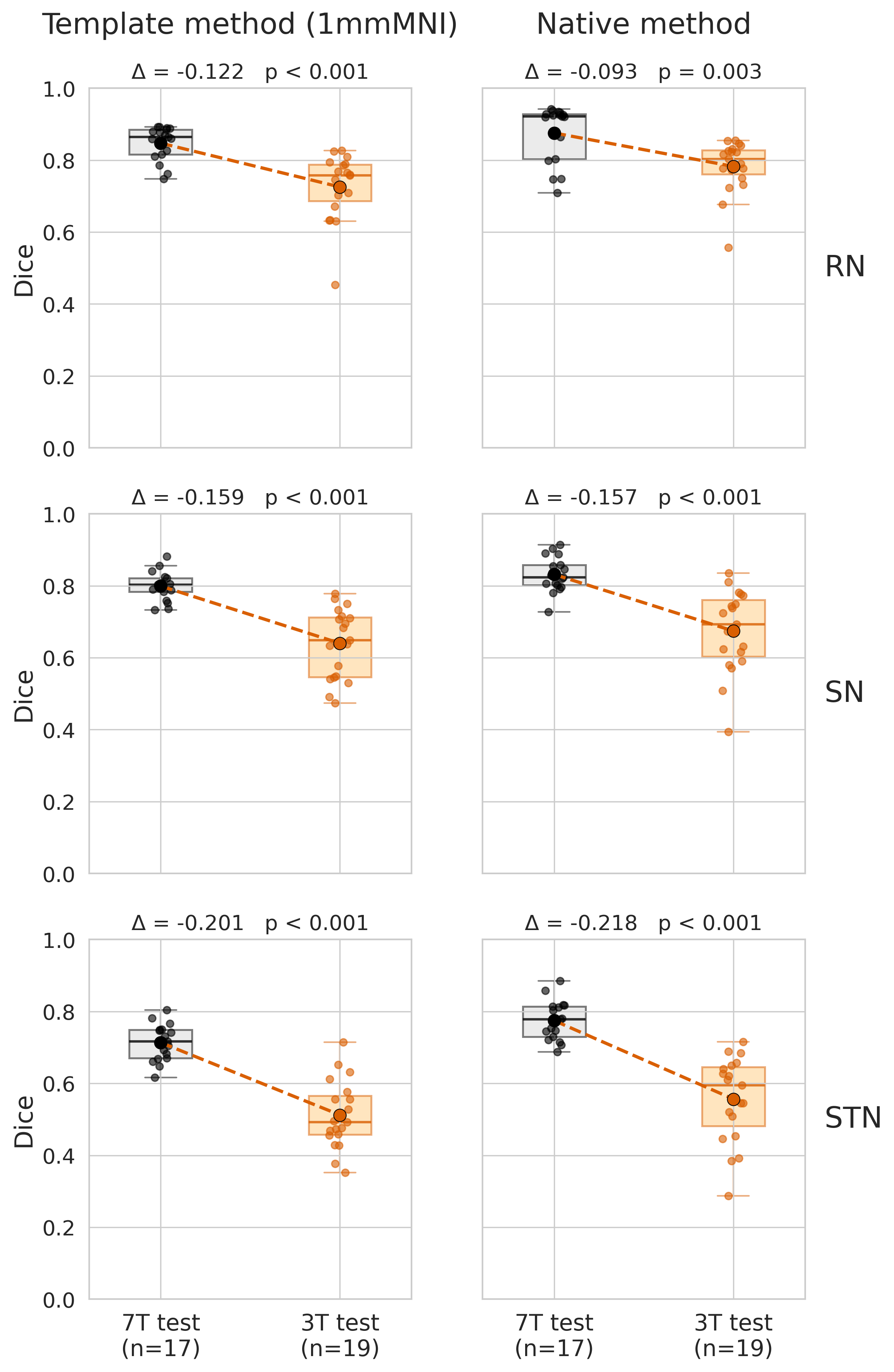}}
\caption{Analysis of knowledge transferability of 7T-only trained models to 3T data. Dice scores comparison between the 7T and 3T test subsets for our two best performing pipelines -- the 1mm variation of the Template method and the Native method -- trained on the 7T train subset.}
\label{fig:q3_plot}
\end{figure}

When trained on 7T only, all models showed a clear drop in accuracy on the 3T test subset relative to 7T, indicating a strong cross-field domain gap (Figure \ref{fig:q3_plot}). This drop is largest for the STN, where the best-performing configuration under 7T-only training (Native pipeline) achieves Dice 0.556 ± 0.119 on 3T, compared to 0.775 ± 0.055 on 7T (Table \ref{tab:q1_q2} vs Table \ref{tab:q4}). RN and SN show the same pattern (RN native: 0.782 ± 0.073 on 3T vs 0.875 ± 0.080 on 7T; SN native: 0.674 ± 0.114 on 3T vs 0.831 ± 0.049 on 7T). The loss is particularly noticeable for the higher resolution version of the Template method, where for the STN, the mean Dice score drops from 0.650 ± 0.166 to 0.353 ± 0.157 (Table \ref{tab:q1_q2} vs Table \ref{tab:q4}). This points towards a poor direct knowledge transferability from high-field MRI images to clinical images.

\begin{figure}[!t]
\centerline{\includegraphics[width=1\columnwidth]{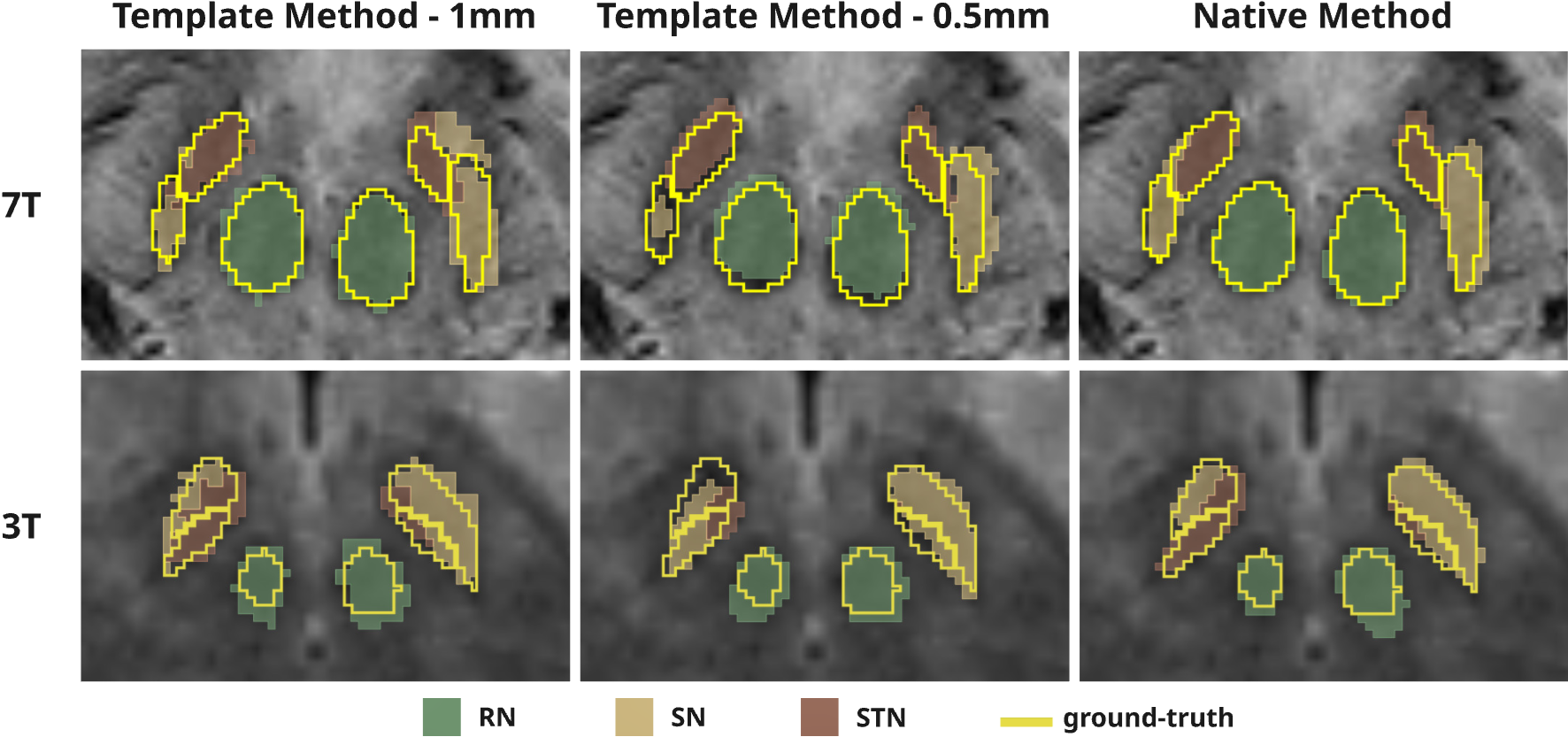}}
\caption{Illustration of the output of each method for one 7T (top) and one 3T (bottom) images of the test subset. The manual segmentation is overlayed (yellow line) on top of the model outputs.}
\label{fig:segmentation_examples}
\end{figure}

Figure \ref{fig:segmentation_examples} illustrates that, on 3T images, the Template method exhibits visible misalignment and boundary distortions after inverse mapping, while the Native method preserves more anatomically plausible localization but remains limited by reduced tissue contrast and boundary visibility.

\subsection{Q4 - Usefulness of synthetic 3T images to support knowledge transferability}
\begin{figure}[!t]
\centerline{\includegraphics[width=1\columnwidth]{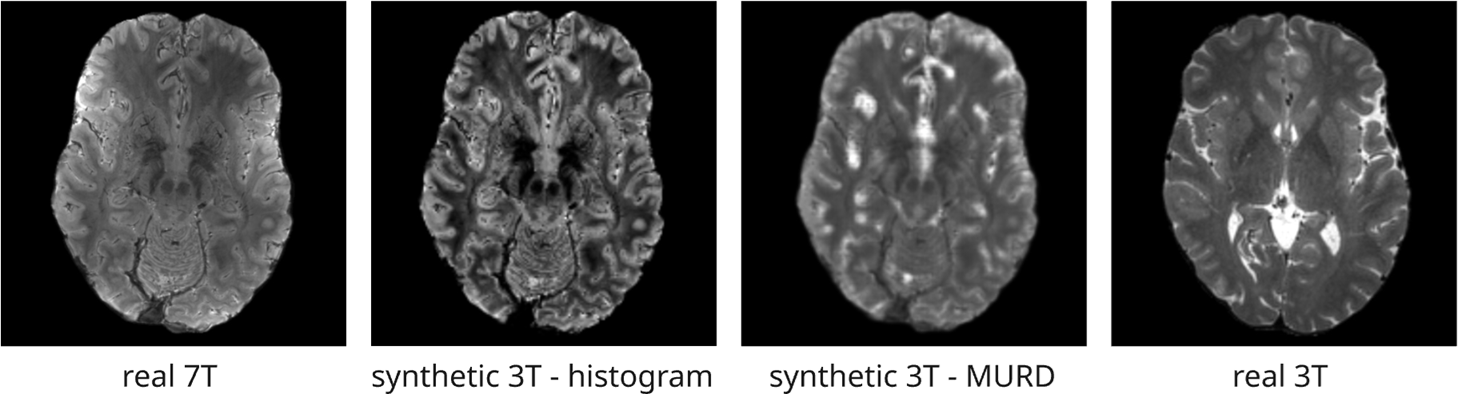}}
\caption{Visual comparison between a slice of a real 7T image of the CBS dataset and the corresponding slice of the synthetically generated 3T images, both by the simple (histogram matching) and the advanced (adapted MURD) methods. A slice of a real 3T image of the Paired dataset (right) is shown for reference.}
\label{fig:synthetic_3T}
\end{figure}

Figure \ref{fig:synthetic_3T} allows the qualitatively comparison of the two synthetic image generation methods used.
MURD-generated synthetic 3T images successfully reproduce the lower contrast and reduced tissue boundary definition, characteristic of real 3T acquisitions, while preserving the overall anatomical content of the 7T source. However, subtle differences in intensity distribution and local texture remain between the synthetic and real 3T images, suggesting that the style transfer is only partially capturing the full complexity of 3T acquisition characteristics. Despite these handicaps, we considered the synthetic images generated by MURD to be of significant enough quality to train our segmentation pipelines.

Synthetic 3T augmentation had method-dependent and structure-dependent effects. Focusing on the two best performing pipelines:
\subsubsection{Template method (1mm MNI)}
On the 3T test set, adding synthetic data improved RN Dice (Template 1mm: 0.725 → 0.751/0.754 for hist/MURD; Table \ref{tab:q4}), and dramatically reduced RN HD95 compared to the 7T-only baseline (from 3.14 ± 3.74 to ~1.1 mm). However, for SN and STN the improvements were smaller and inconsistent, with some configurations showing reduced Dice and/or increased variability.

\subsubsection{Native method}
On the 3T test set, MURD augmentation produced the most consistent (though modest) improvements, particularly for STN (0.556 ± 0.119 → 0.567 ± 0.115) and RN (0.782 ± 0.073 → 0.783 ± 0.068), with limited change for SN (Table \ref{tab:q4}).

\begin{table}
\begin{center}
\caption{Segmentation performance on the 3T test set across training regimes. Results are shown as mean ± standard deviation. Best results per structure are in bold.}
\label{tab:q4}
\begin{tabular}{lllcc}
\toprule
Structure & 
& Training & Dice & HD95 (mm) \\
\midrule
RN & Template (1mm) & 7T only & 0.725 ± 0.091 & 3.14 ± 3.74 \\
RN & Template (1mm) & 7T + synth (hist) & 0.751 ± 0.091 & 1.11 ± 0.26 \\
RN & Template (1mm) & 7T + synth (MURD) & 0.754 ± 0.078 & 1.08 ± 0.19 \\
\rowcolor[gray]{.9}
RN & Template (0.5mm) & 7T only & 0.644 ± 0.089 & 1.59 ± 0.48 \\
\rowcolor[gray]{.9}
RN & Template (0.5mm) & 7T + synth (hist) & 0.472 ± 0.203 & - \\
\rowcolor[gray]{.9}
RN & Template (0.5mm) & 7T + synth (MURD) & 0.525 ± 0.178 & 2.48 ± 1.44 \\
RN & Native & 7T only & 0.782 ± 0.073 & 1.08 ± 0.22 \\
RN & Native & 7T + synth (hist) & 0.749 ± 0.082 & 1.12 ± 0.29 \\
RN & Native & 7T + synth (MURD) & \textbf{0.783 ± 0.068} & \textbf{1.02 ± 0.21} \\
\hline
SN & Template (1mm) & 7T only & 0.640 ± 0.097 & 2.22 ± 2.00 \\
SN & Template (1mm) & 7T + synth (hist) & 0.661 ± 0.094 & 1.68 ± 0.46 \\
SN & Template (1mm) & 7T + synth (MURD) & 0.620 ± 0.126 & 1.94 ± 0.88 \\
\rowcolor[gray]{.9}
SN & Template (0.5mm) & 7T only & 0.638 ± 0.065 & 1.80 ± 0.34 \\
\rowcolor[gray]{.9}
SN & Template (0.5mm) & 7T + synth (hist) & 0.401 ± 0.184 & 2.66 ± 0.77 \\
\rowcolor[gray]{.9}
SN & Template (0.5mm) & 7T + synth (MURD) & 0.434 ± 0.167 & 3.01 ± 1.04 \\
SN & Native & 7T only & 0.674 ± 0.114 & 1.58 ± 0.53 \\
SN & Native & 7T + synth (hist) & 0.649 ± 0.121 & 1.90 ± 0.72 \\
SN & Native & 7T + synth (MURD) & \textbf{0.679 ± 0.112} & \textbf{1.51 ± 0.51} \\
\hline
STN & Template (1mm) & 7T only & 0.512 ± 0.095 & 3.01 ± 2.89 \\
STN & Template (1mm) & 7T + synth (hist) & 0.504 ± 0.110 & 2.28 ± 0.96 \\
STN & Template (1mm) & 7T + synth (MURD) & 0.475 ± 0.126 & 2.45 ± 1.10 \\
\rowcolor[gray]{.9}
STN & Template (0.5mm) & 7T only & 0.353 ± 0.157 & 2.80 ± 1.03 \\
\rowcolor[gray]{.9}
STN & Template (0.5mm) & 7T + synth (hist) & 0.182 ± 0.189 & - \\
\rowcolor[gray]{.9}
STN & Template (0.5mm) & 7T + synth (MURD) & 0.304 ± 0.192 & 3.18 ± 1.08 \\
STN & Native & 7T only & 0.556 ± 0.119 & 2.29 ± 1.33 \\
STN & Native & 7T + synth (hist) & 0.530 ± 0.148 & 2.02 ± 0.85 \\
STN & Native & 7T + synth (MURD) & \textbf{0.567 ± 0.115} & \textbf{1.89 ± 0.81} \\
\bottomrule
\end{tabular}
\end{center}
\end{table}

\begin{figure}[!t]
\centerline{\includegraphics[width=0.6\columnwidth]{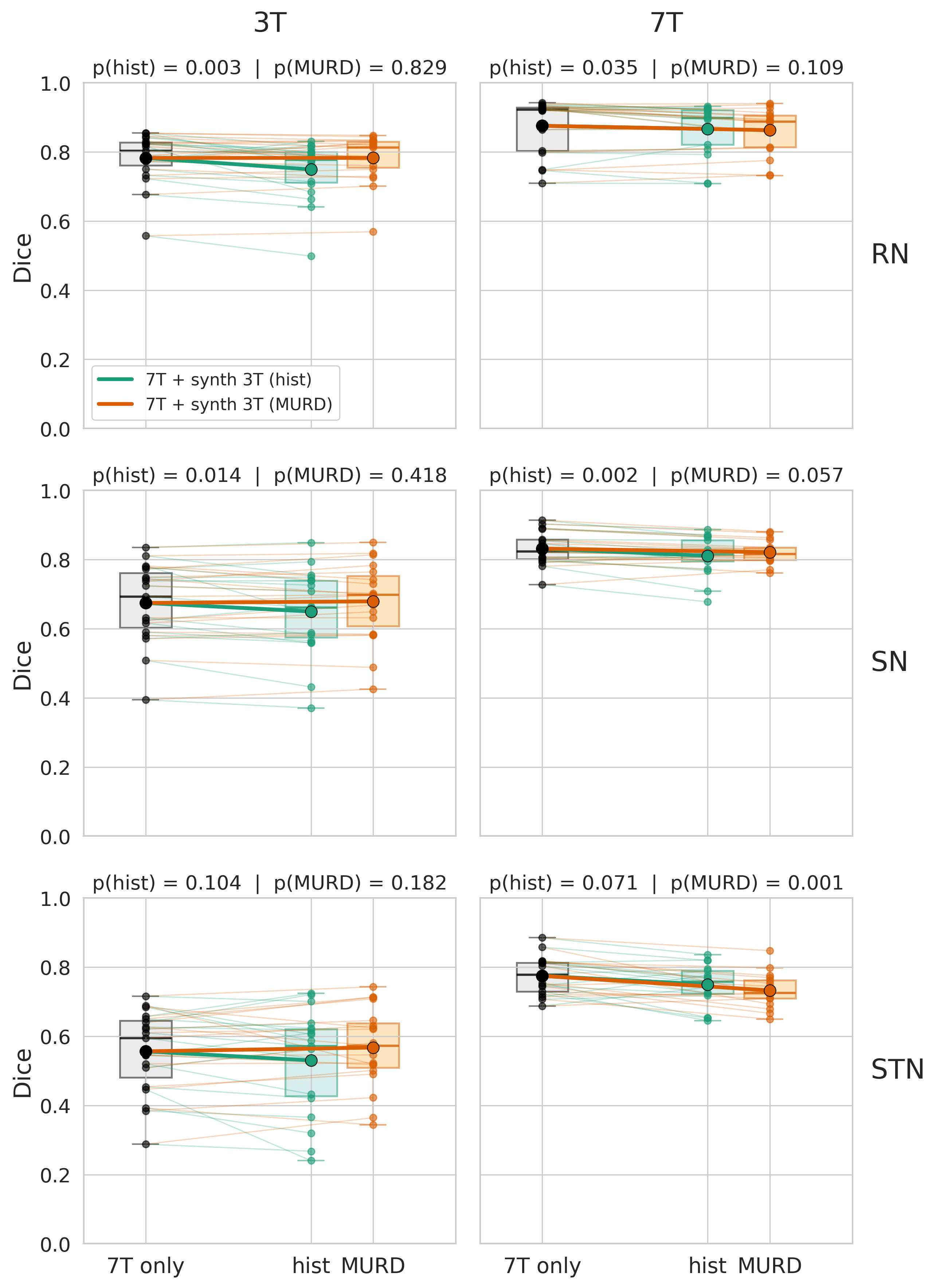}}
\caption{Effect of the addition of synthetic 3T data to training on the Template method. The results using both the simple (histogram matching) and the advanced (adapted MURD) synthetic generation methods are shown separately on the 3T and the 7T test subsets.}
\label{fig:q4_plot}
\end{figure}

Besides understanding the potential benefit that the synthetic images could have on the knowledge transferability to 3T images, it was also important to verify if it would not negatively affect the performance in the 7T images. Figure \ref{fig:q4_plot} shows the Dice score distributions of the best performing method -- Native pipeline -- under the different training regimes, both on the 7T and on the 3T test subsets. Across structures and field-strength, p-values are high, denoting no significant influence of the addition of the synthetic data. While 3T data performance seems to slightly benefit, or at least not be affected by it, the performance on 7T slightly decays when the generated images are added. 

\subsubsection{Additional Volumetric and Surface Area Analysis}
Besides the Dice and HD95 metrics computed, we sought to understand the effects of the different methods and training conditions (inclusion or not of synthetic 3T data) directly on the volume and surface area measures. For this end, we calculated the Volume (V) and Surface Area (SA) relative differences between the predicted segmentations and the ground-truth labels - see the Figures in Appendix \ref{fig:vol_area_method_I_1mm} and \ref{fig:vol_area_method_II_native}.

The Template method exhibited a tendency to overestimate the volume of all three structures in 3T images, with the STN showing the largest and most variable relative differences. In contrast, the Native pipeline demonstrated more centered and symmetric distributions of relative volumetric error for both 3T and 7T subsets, although in this case, the structures tended to be slight underestimated across field-strengths and training conditions. The addition of MURD-generated synthetic 3T data to training caused a reduction in the predicted structures volume and surface area for both methods. This effect was not so pronounced on the 7T subset.



\section{Discussion}

This work was designed around two clinically motivated questions in two areas: (i) whether template normalization pipelines are compatible with the requirement for native-space accuracy in intervention planning (Q1–Q2); and (ii) whether a 7T-trained model that has better learned the target structures' shapes in clearer images can be made to generalize to 3T clinical MRI, potentially via synthetic 3T augmentation, without requiring additional labels generation (Q3–Q4). The results support a consistent narrative: native-space processing is the most reliable route to native-space segmentations, and while synthetic data can help in specific cases, it does not fully bridge the 7T to 3T domain gap.

\subsubsection{Native space segmentation outperforms MNI-based approaches}
Our results indicate that the Native method outperformed the Template method across all three structures and both field strengths. The degradation introduced by the inverse affine transformation from MNI to native space was particularly pronounced for the STN -- the smallest and most irregularly shaped of the three target structures. This finding aligns with the known sensitivity of small structure segmentation to registration errors \cite{DBS_non_linear_deformations} and with the concerns raised by \textcite{variability_1} and \textcite{variability_2} regarding the high inter-subject variability of the STN, which is inherently poorly captured by population-averaged templates. \textcite{DBS_non_linear_deformations} previously showed that nonlinear deformation algorithms vary considerably in their ability to map atlas-based segmentations to individual anatomy, and our results suggest that even in a deep learning context, the transformation back to native space remains a critical bottleneck.

The fact that the 0.5mm MNI variation of the Template method did not improve upon the 1mm version further supports this interpretation. If the accuracy loss were primarily due to insufficient resolution in the template space, higher-resolution templates should have yielded better results. Instead, the persistent degradation points to the affine transformation itself—and potentially the interpolation of binary label maps—as the main source of error. This is consistent with the findings of \textcite{klein14registrations}, who demonstrated that different registration algorithms introduce variable degrees of error in subcortical structure mapping, with smaller nuclei being disproportionately affected.

The superior performance of the Native method can be attributed to the fact that the segmentation model operates directly on the native image geometry, avoiding any spatial distortion of the output labels. Although it still relies on an atlas (CIT168) for ROI localization, this registration is applied only to define a bounding box and does not distort the patient's brain image nor the labels. The CIT168 atlas has been validated for subcortical applications \cite{cit168, cit168_success, MNI_distortion_and_CIT168_success}, and its probabilistic maps provided a reliable initial localization of the target structures across the diverse datasets used in this study.

These results reinforce the argument made by \textcite{variability_3} in favor of patient-specific segmentation, particularly for pre-surgical targeting. In clinical practice, even small localization errors can translate into suboptimal electrode placement and reduced therapeutic outcomes [1, 3]. The avoidance of spatial normalization in the segmentation pipeline thus appears to be a clinically meaningful advantage, while the use of brain templates will remain a facilitator step for large consortia studies merging data from different sources.

\subsubsection{Limited impact of synthetic 3T data augmentation}
The second aim of this study was to explore whether synthetically generated 3T-like images could be used to improve the adaptability of 7T-trained models to clinical 3T inputs. This was motivated by the assumption that segmentation models trained with high-quality 7T data could better learn the anatomy and shape constraints of the small subcortical target structures and apply this knowledge on lower-resolution images. Previous work \cite{wang20257t} showed that this is not directly feasible, so the creation of synthetic 3T images from the original 7T ones aimed at increasing the generalizability of a segmentation model dependent solely on 7T data.

Our results show that the addition of synthetic 3T data generated using an adaptation of MURD \cite{MURD} to training did not produce the anticipated improvement. While slight trends toward better 3T Dice scores were observed -- particularly for the STN in Native method (Table \ref{tab:q4}) -- the overall effect was modest and inconsistent across structures and methods. The comparison between MURD-generated and histogram-matched synthetic data (Figure \ref{fig:q4_plot}) further revealed that the more sophisticated MURD framework offered only marginal gains over simple intensity histogram standardization. This is a somewhat surprising outcome, given that MURD explicitly disentangles content and style representations and has shown success in multi-site harmonization tasks \cite{MURD}.

Several factors may explain this limited effect. First, the domain gap between 7T and 3T extends beyond global intensity differences to include fundamentally different contrast-to-noise ratios, partial volume effects, and tissue boundary visibility—properties that are difficult to replicate through image-level style transfer alone. As noted by Zhao et al. (2019) in the context of synthetic data for medical image segmentation, surface-level appearance matching may be insufficient when the underlying signal characteristics differ substantially. Second, the relatively small number of 3T training images available to MURD (19 images from two datasets) may have limited the model's ability to capture the full variability of 3T acquisition characteristics.

It is worth noting that the synthetic data augmentation did not greatly degrade 7T performance, both for the Template method 1~mm MNI variation and the Template method, which is an important practical consideration. This suggests that mixed-field training is at least safe, even if its benefits for 3T generalization remain uncertain.

\subsubsection{Limitations and future directions}
We identified the following limitations that should be taken into account when interpreting the results: The sample sizes, particularly for the 3T test subset, were small (12 images: 2 Paired + 10 OASIS), limiting the statistical power to detect differences between training conditions. The OASIS 3T images, acquired at 1mm isotropic resolution, represent a particularly challenging test case due to their lower spatial resolution compared to the 7T training data, and the absence of these images from the training set means that the models were evaluated in a fully out-of-distribution setting. This may explain the low Dice scores achieved when compared to other state-of-the-art studies using cross-validation scenarios. Additionally, all spatial transformations in Template-based Method were affine; the use of nonlinear registration could potentially reduce the accuracy loss associated with the MNI-based pipeline, though at the cost of increased computational complexity and the risk of introducing local distortions in small structures, which would again reduce the significance of the work in a clinical scenario.

The exploratory nature of the synthetic data experiments should also be acknowledged at two levels. Firstly, we used UNet for its proven segmentation capabilities in small datasets and in medical context. UNet models still achieve state-of-the-art performance and are used to construct reliable segmentation pipelines \cite{baniasadi2023dbsegment}. Although we expect our findings to hold for other model architectures, future work may explore whether the different model families are affected to the same extent. In particular, it may be interesting to test the generalizability between different field-strengths of resolution-adaptive segmentation frameworks \cite{diaz2025learning, henschel2022fastsurfervinn}

Secondly, the MURD framework was adapted from a multi-site harmonization context to a cross-field-strength translation task, and it is possible that architectures specifically designed for 7T-to-3T synthesis (such as those based on diffusion models or conditional GANs trained with paired data) might yield more effective synthetic training images. Future work could also investigate the use of physics-informed simulation approaches that explicitly model the signal differences between field strengths, rather than relying on learned style transfer.

Finally, the evaluation was limited to the Dice coefficient, HD95 and relative volumetric differences. Clinically relevant metrics such as the assessment of electrode targeting accuracy in a simulated surgical planning scenario could provide a more complete picture of the practical impact of these methods. Since the structures evaluated in this work are closely related to PD, future work should also assess the validity of our findings in PD patients' MRI images, or at least in the same group age as PD patients (>60 years old).

\section{Conclusion}
This study evaluated the effect of space normalization on UNet-based automatic segmentation of the Subthalamic Nucleus and neighboring basal ganglia structures, and explored the use of synthetically generated 3T-like training data to improve model generalization to clinical field strengths.

Our findings support two main conclusions. First, the native space segmentation pipeline consistently outperformed the MNI-based approach, with the STN showing the largest performance gap. The accuracy degradation introduced by inverse affine transformations represents a meaningful limitation of normalization-based workflows, particularly for small and anatomically variable structures.

Second, the use of synthetic 3T images generated with an adapted multi-site harmonization framework (MURD) did not yield the substantial improvement in 3T segmentation performance that was hypothesized. Although the MURD-based approach showed marginally better results than simple histogram matching, neither strategy produced a statistically robust enhancement.

Although the generalizability of our findings remains to be tested for different MRI field-strengths and clinically relevant brain structures, our results suggest a clear benefit of the native space processing within the tested conditions.

\section*{Funding Details}
The study was supported by the Johannes Amos Comenius project Brain Dynamics, No\\
CZ.02.01.01/00/22\_008/0004643.

\section*{Acknowledgments}
The access to the computational infrastructure of the OP VVV funded project\\ CZ.02.1.01/0.0/0.0/16\_019/0000765 ``Research Center for Informatics'' is also gratefully acknowledged.

%
%
%
%
\printbibliography{}

\newpage
\appendix
\renewcommand{\thefigure}{A\arabic{figure}}

\setcounter{figure}{0}
\section*{Appendix}

\begin{figure}[H]
\centerline{\includegraphics[width=0.6\columnwidth]{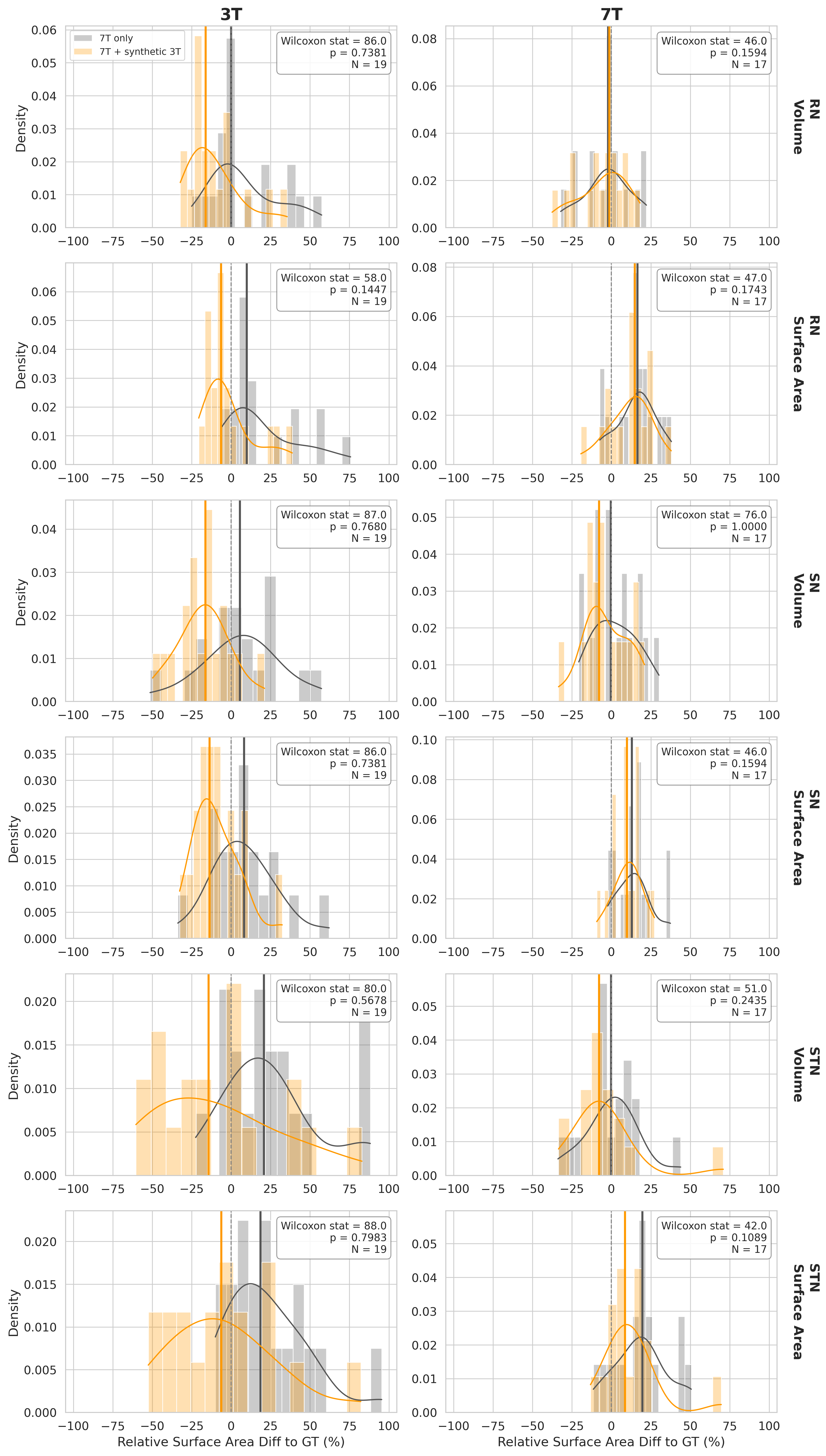}}
\caption{\textbf{Template method (1~mm): }Statistical distribution of the relative differences between the predicted segmentations of Template method 1mm MNI variation and the ground-truth labels. The results are shown separately for each structure and for the 3T and 7T subsets. For each structure/field pair, both the results of Template method trained only with 7T and with 7T + 3T synthetic data (MURD generated) are shown.}
\label{fig:vol_area_method_I_1mm}
\end{figure}

\begin{figure}[!t]
\centerline{\includegraphics[width=0.6\columnwidth]{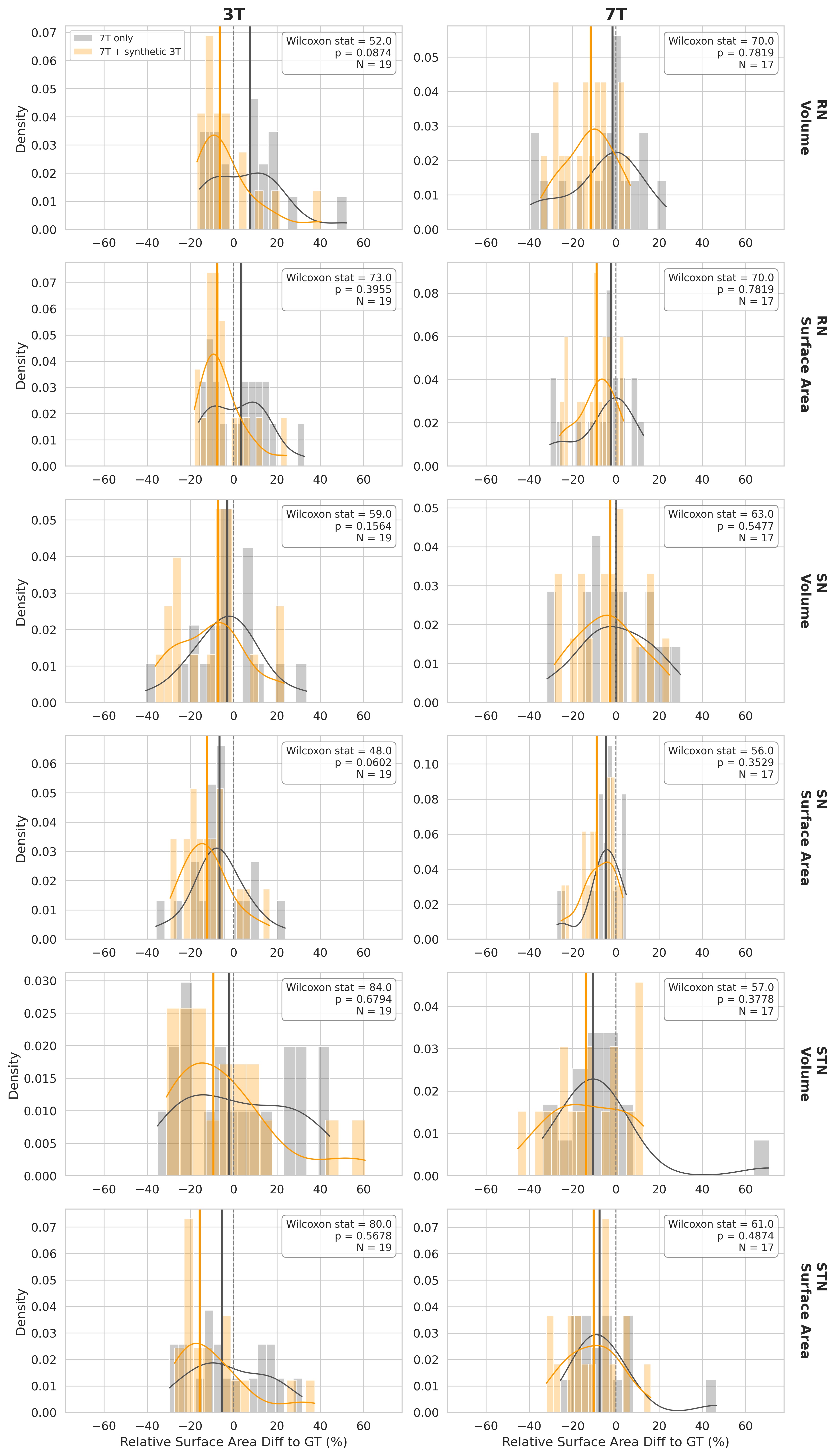}}
\caption{\textbf{Native method}: Statistical distribution of the relative differences between the predicted segmentations of Native method and the ground-truth labels. The results are shown separately for each structure and for the 3T and 7T subsets. For each structure/field pair, both the results of Native method trained only with 7T and with 7T + 3T synthetic data (MURD generated) are shown.}
\label{fig:vol_area_method_II_native}
\end{figure}

\end{document}